\documentclass[10pt,a4paper]{article}
\usepackage[T1]{fontenc}
\usepackage{lmodern}
\usepackage[margin=22mm,headheight=14pt]{geometry}
\usepackage{amsmath,amssymb,booktabs,tabularx,array,multirow,fix-cm}
\usepackage{tikz}
\usetikzlibrary{arrows.meta,positioning,fit,calc,shapes.geometric,backgrounds}
\usepackage{xcolor,enumitem,fancyhdr,caption,microtype,titlesec}
\usepackage[unicode,colorlinks=true,linkcolor=blue!45!black,citecolor=blue!45!black,urlcolor=blue!45!black]{hyperref}

\definecolor{fmInk}{HTML}{222222}
\definecolor{fmGray}{HTML}{6B7280}
\definecolor{fmBlue}{HTML}{0072B2}
\definecolor{fmBlueBg}{HTML}{EAF3FA}
\definecolor{fmGrayBg}{HTML}{F3F4F6}
\definecolor{fmOrange}{HTML}{D55E00}
\definecolor{fmOrangeBg}{HTML}{FCEFE6}
\definecolor{fmGreen}{HTML}{009E73}
\definecolor{fmGreenBg}{HTML}{E6F5F0}

\setlist{nosep,leftmargin=1.8em}
\titlespacing*{\section}{0pt}{1.1ex plus 0.2ex minus .2ex}{0.5ex}
\titlespacing*{\subsection}{0pt}{0.8ex plus 0.2ex minus .2ex}{0.35ex}
\newcommand{\tool}[1]{\texttt{\small\detokenize{#1}}}
\newcolumntype{Y}{>{\raggedright\arraybackslash}X}
\newcolumntype{L}[1]{>{\raggedright\arraybackslash}p{#1}}

\tikzset{
  every picture/.style={font=\fontsize{8}{10}\selectfont,text=fmInk},
  box/.style={draw=fmInk,line width=.55pt,rounded corners=2pt,align=center,inner sep=3.2pt,minimum height=7.5mm},
  agentbox/.style={box,draw=fmBlue,line width=.85pt,fill=fmBlueBg},
  databox/.style={box,fill=fmGrayBg},
  gate/.style={box,draw=fmOrange,line width=.75pt,fill=fmOrangeBg},
  okbox/.style={box,draw=fmGreen,line width=.75pt,fill=fmGreenBg},
  edge/.style={-{Stealth[length=3.8pt,width=2.8pt]},line width=.6pt},
  interact/.style={{Stealth[length=3.8pt,width=2.8pt]}-{Stealth[length=3.8pt,width=2.8pt]},line width=.6pt},
  feedback/.style={edge,dashed,draw=fmGray},
  lab/.style={font=\fontsize{7}{8.5}\selectfont,fill=white,inner sep=1.2pt},
  layer/.style={font=\fontsize{7.5}{9}\selectfont\bfseries,text=fmGray,rotate=90,anchor=south}
}

\title{\vspace{-1.2em}\bfseries Megrez: A Knowledge-Guided Agent for Formal Mathematics\\over Morse--Kelley Set Theory\vspace{-0.4em}}
\author{Tiancheng Xu\textsuperscript{1}\qquad Wensheng Yu\textsuperscript{1}}
\hypersetup{pdfauthor={Tiancheng Xu, Wensheng Yu}}
\date{\vspace{-1.2em}}

\begin{document}
\maketitle
\footnotetext[1]{School of Electronic Engineering, Beijing University of Posts and Telecommunications, Beijing 100876, China.\par
\noindent Email: \href{mailto:xutiancheng@bupt.edu.cn}{\texttt{xutiancheng@bupt.edu.cn}} (Tiancheng Xu).}

\begin{abstract}
General-purpose coding agents equipped with proof-assistant tools have shown strong performance on standalone competition problems, yet cumulative textbook formalization inside a low-resource mathematical library still faces three obstacles: aligning textbook concepts with library declarations, keeping long-horizon search synchronized with live proof states, and integrating locally verified proofs reliably into a growing codebase. We present Megrez, a knowledge-guided formalization agent for Rocq libraries built on Morse--Kelley (MK) set theory. Megrez coordinates a general coding agent through three mechanisms: (1) a source-rebuildable knowledge graph merging formal declarations, compiler references (\tool{.glob}), and curated metadata to support retrieval via Chinese aliases, concept clusters, proof dependencies, and multi-lane full-text ranking; (2) a feedback-driven 1--5 tactic short-burst proving loop paired with a goal-local branch ledger and exponential-interval structured reflection; and (3) a three-check acceptance pipeline (statement fidelity, axiom auditing, and full builds) coupled with a multi-corpus Provider stack and a candidate-bound recoverable batch state machine. Deployment and evaluation on a point-set topology corpus demonstrate effective cross-chapter formalization and scalable premise retrieval.
\end{abstract}
\noindent\textbf{Keywords:} formal mathematics; Morse--Kelley set theory; Rocq; general coding agents; knowledge graph; interactive theorem proving

%======================================================================
\section{Introduction}
%======================================================================
Formal mathematics represents mathematical definitions, theorems, and proofs as objects that a proof assistant can check mechanically. In Rocq, users construct proofs interactively with tactics and the kernel type-checks the resulting proof term~\cite{rocq}. While kernel verification guarantees logical soundness, formalizing a textbook section by section inside a maintainable library remains labor-intensive: one must align informal terminology with existing declarations, discharge set-theoretic side conditions left tacit on paper, and ensure that new developments integrate cleanly with the surrounding codebase.

Large language models (LLMs) are advancing proof automation along two main lines: training specialized provers via synthetic data and reinforcement learning for whole-proof generation~\cite{deepseekprover2,kimina,goedel2}, and conditioning step-by-step tactic generation on proof states~\cite{gptf,leandojo,copra}. Recently, Numina-Lean-Agent demonstrated a more general paradigm: coupling a general-purpose coding agent with Lean tools exposed through the Model Context Protocol (MCP)~\cite{mcp} achieves competitive results on both competition benchmarks and research-level formalization without training a task-specific model~\cite{numina}. Megrez adopts this general-agent-plus-tools paradigm for Rocq libraries built over Morse--Kelley (MK) axiomatic set theory~\cite{kelley,morse,sunyu}, targeting \textbf{cumulative textbook formalization} across three core challenges:
\begin{enumerate}
  \item \textbf{Concept Alignment:} MK set-theory libraries are much smaller than Mathlib and rarely appear in pretraining data. A single textbook concept may correspond to a definition, abbreviation, or numbered theorem, and set-theoretic encoding requires explicit side conditions (e.g., proving that a class is a set \tool{Ensemble} rather than a proper class, or that an argument lies in a domain). Identifier string search alone rarely bridges textbook intent to formal infrastructure.
  \item \textbf{State Synchronization:} Mathematically sound plans often fail locally due to hypothesis shapes, notation bindings, or higher-order unification direction. Generating long scripts without intermediate feedback builds on broken states; conversely, repeated local failures usually signal a flaw in the mathematical route or representation rather than a single tactic.
  \item \textbf{Library Integration:} Textbook formalization is cumulative. A locally accepted script must still be checked to confirm that it proves the exact reviewed statement, introduces no unapproved axioms or transitive dependencies on other \tool{Admitted} placeholders, compiles across the full project, and re-enters the retrieval index; multi-theorem batches further require dependency scheduling, concurrency control, and crash recovery.
\end{enumerate}

To address these challenges, we design and implement Megrez (Figure~\ref{fig:architecture}). Our main contributions are:
\begin{itemize}
  \item \textbf{A source-rebuildable formal knowledge graph and multi-lane retrieval (Section~\ref{sec:kg}):} We combine Rocq sources, compiler-resolved \tool{.glob} references, and human-reviewed metadata into a SQLite graph indexing Chinese aliases, concept clusters, proof dependencies, and textbook mappings, supporting scope-before-cap filtering and seven-column FTS5 + Reciprocal Rank Fusion (RRF) ranking.
  \item \textbf{Short-burst interactive proving with goal-local reflection (Section~\ref{sec:prove}):} We pair 1--5 tactic bursts with a per-goal branch ledger, structured reflection at exponential branch checkpoints $\{2,4,8,16,32\}$, and a three-tier knowledge triage discipline (capping top-level skills at $\le 150$ lines while discarding noisy cross-theorem execution trajectories).
  \item \textbf{Three-check acceptance, Provider stacks, and recoverable batches (Sections~\ref{sec:overview} and~\ref{sec:batch}):} We organize formalization around two human review gates, a single-writable-Workspace Provider stack, and a concurrent batch scheduler featuring proof-stripped context fingerprints, canonical proof-hash binding, and four-way deterministic crash recovery.
  \item \textbf{Deployment, scaling benchmarks, and collaborative case studies (Sections~\ref{sec:eval} and~\ref{sec:case}):} We deploy Megrez on MKTopos, a Rocq formalization of Xiong's \emph{Lectures on Point-Set Topology}~\cite{xiong} comprising 262 files, 3,894 declarations, 2,529 completed \tool{Qed} proofs (102,122 lines of proof script), and 60,982 reference edges. We evaluate retrieval latency up to 100,000 nodes ($58.8\times$ exact-lookup speedup, bounded 3--7 SQL queries per call), verify system invariants across 319 automated contract tests, and present three collaborative case studies spanning Tukey's lemma, connectedness of finite and arbitrary product spaces, and the 2D Brouwer fixed-point and Jordan separation theorems.
\end{itemize}

%======================================================================
\section{System Overview and Formalization Lifecycle}\label{sec:overview}
%======================================================================
\subsection{Architecture and Division of Responsibilities}
\begin{figure}[htbp]
\centering
\begin{tikzpicture}[scale=0.92,transform shape]
\node[gate,text width=140mm] (human) at (0,0)
  {\textbf{Mathematician}\quad Textbook / informal statements $\cdot$ Statement review (Gate 1) \& annotation review (Gate 2) $\cdot$ Key decisions};
\node[agentbox,text width=38mm,fill=white] (f) at (-4.75,-1.85) {\tool{mk-formalize}\\Concept inventory \& drafting};
\node[agentbox,text width=38mm,fill=white] (p) at (0,-1.85) {\tool{mk-prove} + \tool{rocq-prove}\\Retrieval, bursts, acceptance};
\node[agentbox,text width=38mm,fill=white] (u) at (4.75,-1.85) {\tool{mk-kg-update}\\KG rebuild \& curation};
\begin{scope}[on background layer]
\node[agentbox,fit=(f)(p)(u),inner xsep=3.5mm,inner ysep=2.5mm,yshift=2.0mm,label={[anchor=north,yshift=-0.3mm,font=\fontsize{8.5}{10}\selectfont\bfseries]north:General Coding Agent (Codex Main Session) + Megrez Workflow Skills}] (agent) {};
\end{scope}
\node[box,text width=40mm] (kg) at (-4.75,-4.05) {\textbf{mk-kg: Knowledge}\\Search / deps / text / packet};
\node[box,text width=40mm] (batch) at (0,-4.05) {\textbf{mk-kg: Batch State}\\DAG schedule / bind / recover};
\node[box,text width=40mm] (rocq) at (4.75,-4.05) {\textbf{rocq-mcp} (external)\\Interact / verify / axioms};
\node[databox,text width=21mm,minimum height=7mm] (dep) at (-6.0,-6.0) {Read-only\\Providers};
\node[databox,text width=21mm,minimum height=7mm,draw=fmBlue,line width=.8pt] (ws) at (-3.15,-6.0) {Writable\\Workspace};
\node[databox,text width=20mm,minimum height=7mm] (kgdb) at (-0.15,-6.0) {KG SQLite\\(rebuildable)};
\begin{scope}[on background layer]
\node[databox,fit=(dep)(ws)(kgdb),inner sep=2.2mm,yshift=1.9mm,label={[anchor=north,yshift=-0.2mm,font=\fontsize{8}{9.5}\selectfont\bfseries]north:Provider Corpus Stack \& Metadata}] (prov) {};
\end{scope}
\node[databox,text width=40mm,double,double distance=1pt] (kernel) at (4.75,-5.85) {\textbf{Rocq Kernel \& Build}\\Proof checking / project build};
\draw[interact] (human) -- node[lab,right]{Tasks \& review} (agent.north);
\draw[interact] (f.south |- agent.south) -- node[lab,left]{Query} (kg.north);
\draw[interact] (agent.south) -- node[lab,right]{Schedule \& auth} (batch.north);
\draw[interact] (u.south |- agent.south) -- node[lab,right]{Tactics \& state} (rocq.north);
\draw[edge] (prov.north -| kg.south) -- node[lab,left]{Extract} (kg.south);
\draw[feedback] (batch.south) -- node[lab,right]{Serial write by main session} (prov.north -| batch.south);
\draw[interact] (rocq.south) -- node[lab,right]{Execute \& audit} (kernel.north);
\draw[edge] (prov.east |- kernel.west) -- node[lab,above]{Build} (kernel.west);
\node[layer] at (-7.55,0) {Human};
\node[layer] at (-7.55,-1.9) {Control};
\node[layer] at (-7.55,-4.05) {Tools};
\node[layer] at (-7.55,-5.9) {Data / Exec};
\end{tikzpicture}
\caption{Architecture of Megrez. Solid arrows denote queries or execution; the dashed arrow indicates serial source writes performed by the main session according to batch state.}
\label{fig:architecture}
\end{figure}
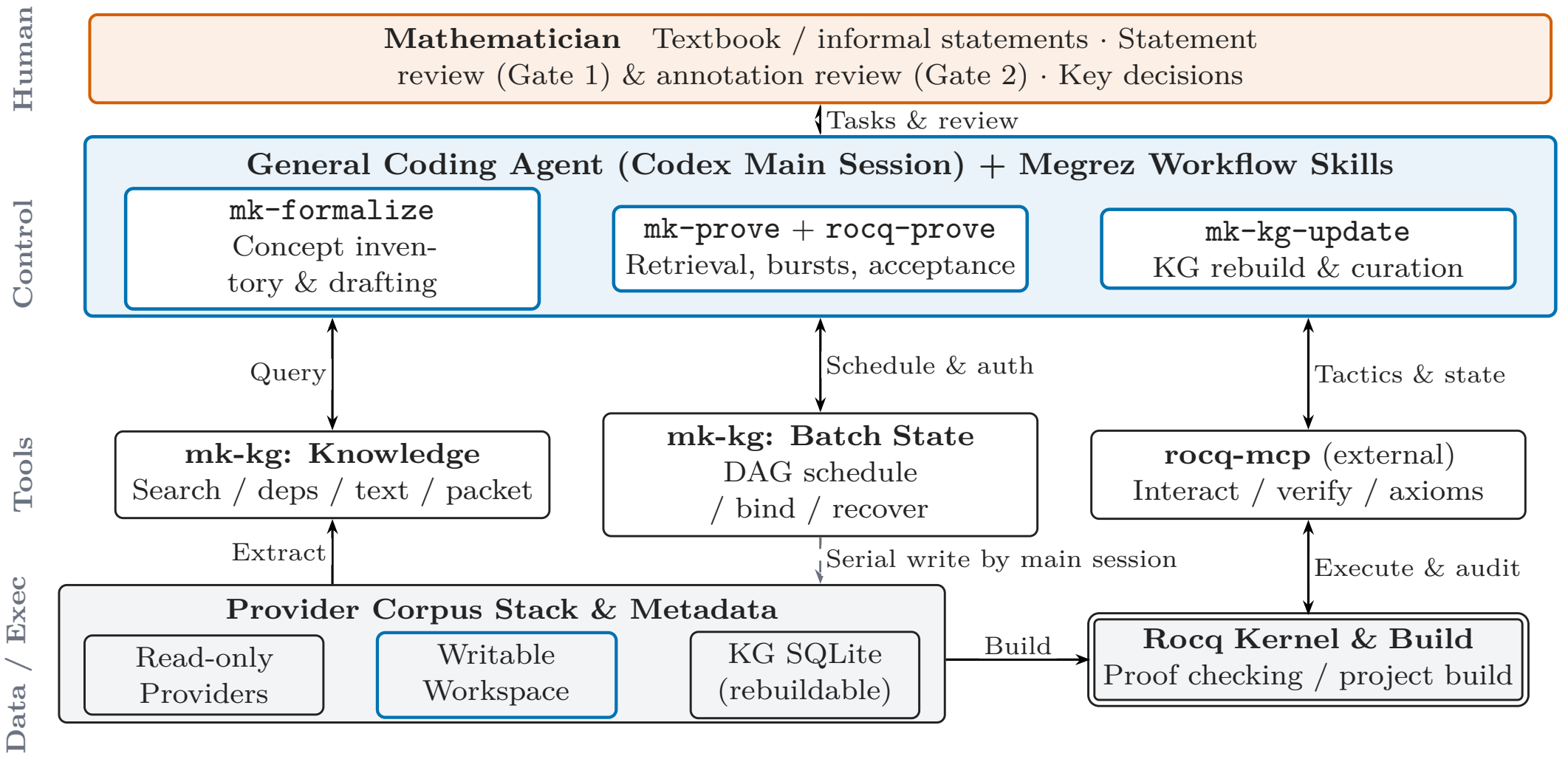

Megrez follows three principles: \emph{traceable knowledge}, \emph{feedback-grounded decisions}, and \emph{verified sedimentation}. As shown in Figure~\ref{fig:architecture}, the system is organized into four layers: (1) the \textbf{Human layer}, where a mathematician supplies textbook statements and reviews formal statements (Gate~1) and semantic annotations (Gate~2); (2) the \textbf{Control layer}, where a general coding agent (currently Codex) loaded with three workflow skills (\tool{mk-formalize}, \tool{mk-prove}/\tool{rocq-prove}, \tool{mk-kg-update}) plans and coordinates tools; (3) the \textbf{Tool layer}, comprising the built-in \tool{mk-kg} MCP server (graph queries, rebuilds, and batch state machine) and an external \tool{rocq-mcp} server (interactive tactic execution, statement verification, and axiom auditing); and (4) the \textbf{Data and Execution layer}, consisting of a Provider dependency stack with a single writable Workspace, an idempotently rebuildable SQLite database, and the Rocq compiler and kernel. Workflow skills define behavioral contracts for the agent, whereas hard safety invariants---capability authentication, compare-and-swap revisions, and candidate proof-hash binding---are enforced deterministically in \tool{mk-kg} tool code.

\subsection{Closed-Loop Lifecycle and Three-Check Acceptance Criterion}\label{sec:accept}
\begin{figure}[htbp]
\centering
\begin{tikzpicture}[node distance=4mm and 5mm,scale=0.92,transform shape]
\node[databox,text width=20mm] (in) {Textbook item /\\informal thm};
\node[agentbox,text width=27mm,right=of in] (form) {\textbf{mk-formalize}\\Inventory $\cdot$ Draft};
\node[gate,text width=20mm,right=of form] (g1) {\textbf{Gate 1}\\Statement fidelity};
\node[agentbox,text width=28mm,right=of g1] (prove) {\textbf{mk-prove}\\Retrieve $\cdot$ Burst $\cdot$ Reflect};
\node[okbox,text width=28mm,below=8.5mm of prove] (acc) {\textbf{3-Check Accept}\\Stmt $\cdot$ Axioms $\cdot$ Build};
\node[agentbox,text width=27mm] (upd) at (acc -| g1) {\textbf{mk-kg-update}\\KG rebuild $\cdot$ Draft};
\node[gate,text width=20mm] (g2) at (acc -| form) {\textbf{Gate 2}\\Concepts \& aliases};
\node[databox,text width=20mm,shape=cylinder,shape border rotate=90,aspect=.18,minimum height=11mm] (kg) at (acc -| in) {Knowledge\\Graph};
\draw[edge] (in) -- (form);
\draw[edge] (form) -- (g1);
\draw[edge] (g1) -- (prove);
\draw[edge] (prove) -- node[lab,right]{Candidate} (acc);
\draw[edge] (acc) -- node[lab,above]{Write} (upd);
\draw[edge] (upd) -- (g2);
\draw[edge] (g2) -- (kg);
\coordinate (kbus) at ($(kg.north)+(0,3.8mm)$);
\draw[feedback] (kg.north) -- (kbus) -| (form.south);
\draw[feedback] (kbus) -| node[lab,pos=.22,above]{Retrieve} ($(prove.south)+(-6mm,0)$);
\draw[feedback] (acc.east) -- ++(3.5mm,0) |- node[lab,pos=.25,right]{Reject} (prove.east);
\end{tikzpicture}
\caption{Formalization lifecycle bracketed by two human review gates.}
\label{fig:lifecycle}
\end{figure}
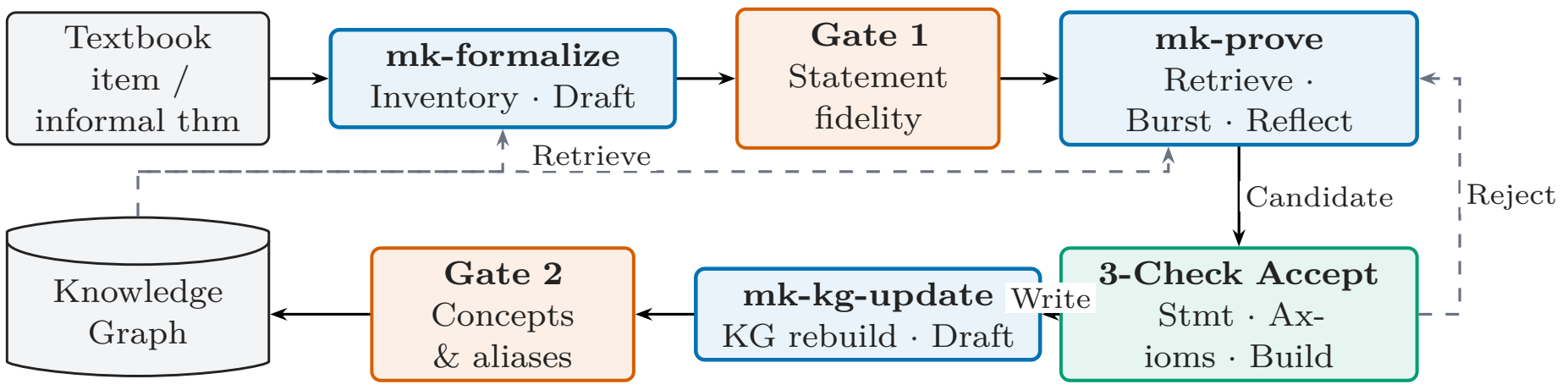

Figure~\ref{fig:lifecycle} illustrates the formalization lifecycle. \tool{mk-formalize} inventories textbook concepts and drafts Rocq statements ending with \tool{Admitted}; after \textbf{Gate~1} human review confirms statement fidelity, \tool{mk-prove} autonomously coordinates retrieval, interactive proving, and verification; once the project builds, \tool{mk-kg-update} rebuilds the graph and drafts concept and Chinese-alias annotations for \textbf{Gate~2} approval. Between the two gates, stage handoffs and crash recovery proceed autonomously within the authorized scope.

Let $\varphi$ be the reviewed statement, $\Gamma$ the current Rocq environment, $p$ a candidate proof term, $\mathcal A$ the axiom whitelist declared by the Workspace Provider, and $\Gamma[p]$ the project state after inserting $p$. A candidate proof is accepted if and only if all three checks pass:
\begin{equation}
  \Gamma\vdash p:\varphi\ \text{(\tool{rocq_verify})},\qquad
  \operatorname{Assumptions}(p)\subseteq\mathcal A\ \text{(\tool{rocq_assumptions})},\qquad
  \operatorname{Build}(\Gamma[p])=\mathrm{pass}\ \text{(full \tool{make})}.
  \label{eq:accept}
\end{equation}
Each check prevents a distinct failure mode: \tool{rocq_verify} prevents the model from weakening the conclusion or strengthening hypotheses; \tool{rocq_assumptions} audits transitive dependencies so $p$ cannot silently route through another unproven \tool{Admitted} item in the same file (which a single-file compile would not catch); and full \tool{make} ensures project-wide compatibility and fresh \tool{.glob} references.

%======================================================================
\section{Knowledge Graph: Bridging Textbook Language and Formal Libraries}\label{sec:kg}
%======================================================================
\subsection{Graph Representation and Idempotent Construction}
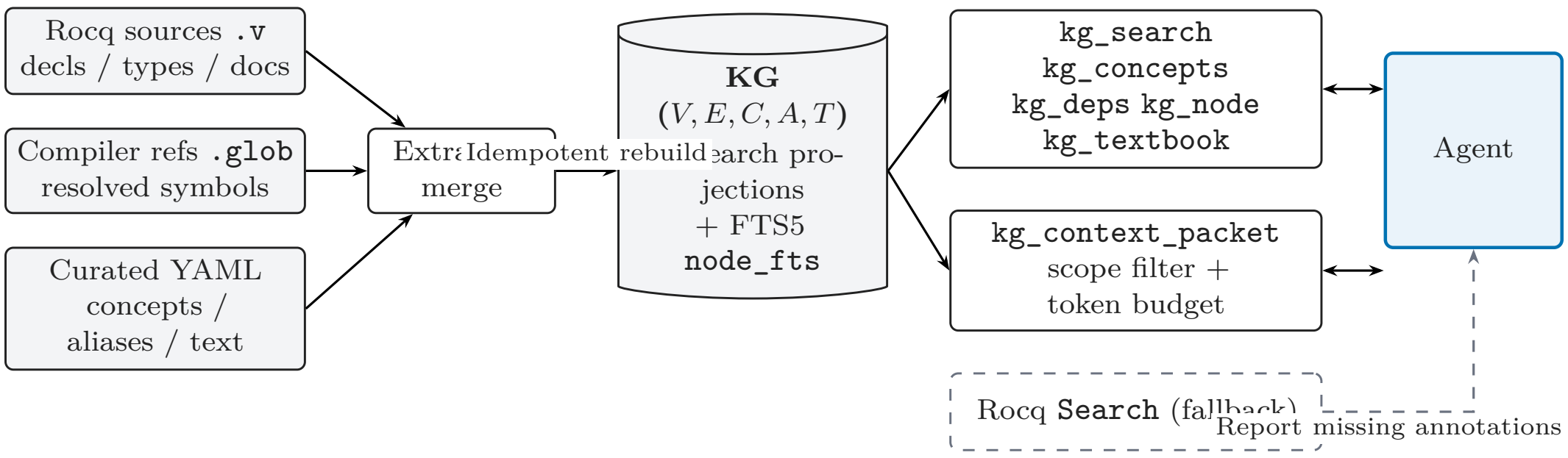
\begin{figure}[htbp]
\centering
\begin{tikzpicture}[node distance=3.2mm and 6mm,scale=0.92,transform shape]
\node[databox,text width=27mm] (v) {Rocq sources \tool{.v}\\decls / types / docs};
\node[databox,text width=27mm,below=of v] (g) {Compiler refs \tool{.glob}\\resolved symbols};
\node[databox,text width=27mm,below=of g] (m) {Curated YAML\\concepts / aliases / text};
\node[box,text width=16mm,right=of g] (ex) {Extract \&\\merge};
\node[databox,shape=cylinder,shape border rotate=90,aspect=.15,text width=24mm,minimum height=23mm,right=of ex,align=center] (db)
  {\textbf{KG ($V,E,C,A,T$)}\\[1pt]+ search projections\\+ FTS5 \tool{node_fts}};
\node[box,text width=34mm,right=6mm of db,yshift=8mm] (q1) {\tool{kg_search}\ \tool{kg_concepts}\\\tool{kg_deps}\ \tool{kg_node}\ \tool{kg_textbook}};
\node[box,text width=34mm,below=4mm of q1] (q2) {\tool{kg_context_packet}\\scope filter + token budget};
\node[agentbox,text width=15mm,right=6mm of q1,yshift=-6mm,minimum height=19mm] (ag) {Agent};
\node[box,dashed,draw=fmGray,text width=34mm,below=4mm of q2] (fb) {Rocq \tool{Search} (fallback)};
\draw[edge] (v.east) -- (ex);
\draw[edge] (g) -- (ex);
\draw[edge] (m.east) -- (ex);
\draw[edge] (ex) -- node[lab,above]{Idempotent rebuild} (db);
\draw[edge] (db.east) -- (q1.west);
\draw[edge] (db.east) -- (q2.west);
\draw[interact] (q1.east) -- (ag.west |- q1.east);
\draw[interact] (q2.east) -- (ag.west |- q2.east);
\draw[feedback] (fb.east) -| node[lab,pos=.22,below]{Report missing annotations} (ag.south);
\end{tikzpicture}
\caption{Knowledge graph construction and KG-first retrieval discipline.}
\label{fig:kg}
\end{figure}

Megrez represents library knowledge as a five-tuple $\mathcal K=(V,E,C,A,T)$ stored in a single SQLite database (schema~v2):
\begin{itemize}
  \item \textbf{Declaration nodes ($V$):} canonical ledger IDs (`\texttt{corpus:file:kind:name}`), formal statements, source spans, comments, and proof status (\tool{qed}/\tool{admitted}/\tool{none}).
  \item \textbf{Reference edges ($E$):} typed relations including \tool{proof_uses}, \tool{statement_mentions}, \tool{uses_notation}, \tool{uses_tactic}, \tool{file_requires}, and \tool{dual_of}. The extractor resolves references primarily from compiler-generated \tool{.glob} files, falling back to lexical matching (tagged with source \tool{token}) only when \tool{.glob} data is missing.
  \item \textbf{Curated semantic layers ($C,A,T$):} concept clusters $C$ group declarations on the same topic by role (\tool{def}/\tool{intro}/\tool{key}/\tool{helper}) and attach reviewed \tool{usage_hints}; aliases $A$ map Chinese textbook terms (e.g., ``nest'', ``finite character'', ``connected component'') to formal identifiers or concepts; textbook mappings $T$ track the formal counterpart and status of each numbered textbook item.
\end{itemize}
As shown in Figure~\ref{fig:kg}, the builder merges \tool{.v} sources, \tool{.glob} references, and version-controlled YAML metadata into deterministic search projections (\tool{search_documents}, \tool{search_terms}) and a seven-column FTS5 index (\tool{node_fts}), computes a path- and timestamp-independent content hash (\tool{kg_content_hash}), and retains a \tool{.previous} snapshot for atomic rollback.

\subsection{KG-First Retrieval and Multi-Lane Fusion}
Megrez enforces a KG-first retrieval sequence: (1) \tool{kg_search} by identifier, Chinese alias, or statement text; (2) \tool{kg_concepts} to inspect cluster members and \tool{usage_hints}; (3) \tool{kg_deps} to inspect the \tool{proof_uses} neighborhood of proved theorems on the same topic as a structural template; and (4) \tool{kg_node} to check exact types and dual theorems. Exploratory Rocq \tool{Search} is permitted only when the graph misses, and any fallback hit must be reported as a missing annotation.

Retrieval is dispatched through versioned profiles. While \tool{legacy-search-v1} and \tool{legacy-premise-v1} preserve fixed-tier compatibility, the schema~v2 profiles (\tool{fts-interactive-v1} and \tool{fts-premise-v1}) pin exact identifier and node-alias matches at the top and fuse prefix, concept-cluster, graph-neighborhood, and seven-column FTS5 lanes (including a symbol lane that maps operators such as \tool{/\ }, \tool{->}, and \tool{:=} into safe tokens, plus a Chinese character bigram lane) using multi-column BM25 and Reciprocal Rank Fusion ($\operatorname{RRF}(v)=\sum_{\ell}(60+\operatorname{rank}_{\ell}(v))^{-1}$). This eliminates both full-table SQL \tool{LIKE} substring scans and per-hit $N+1$ concept queries.

\subsection{Budgeted Premise Packets and Scope-Before-Cap Filtering}
To assemble a bounded context for proof search or read-only workers, \tool{kg_context_packet} packs ranked premises greedily within total budget $B$ and premise-channel budget $B_P$, enforcing $\widehat\tau(\operatorname{Packet}(P))\leq B$ and $\sum_{v\in P}\widehat\tau(v)\leq B_P$ under a deterministic token estimator $\widehat\tau$. When a caller supplies a candidate scope snapshot $S$ (\tool{CandidateScopeSnapshot}), the engine materializes a request-private \tool{TEMP eligible_nodes} table and joins it before per-lane caps so that $P\subseteq S$, while physically masking outgoing \tool{proof_uses} edges of the target theorem itself to prevent benchmark leakage.

%======================================================================
\section{Feedback-Driven Interactive Proving}\label{sec:prove}
%======================================================================
\subsection{Short-Burst Loop and Goal-Local Branch Ledger}
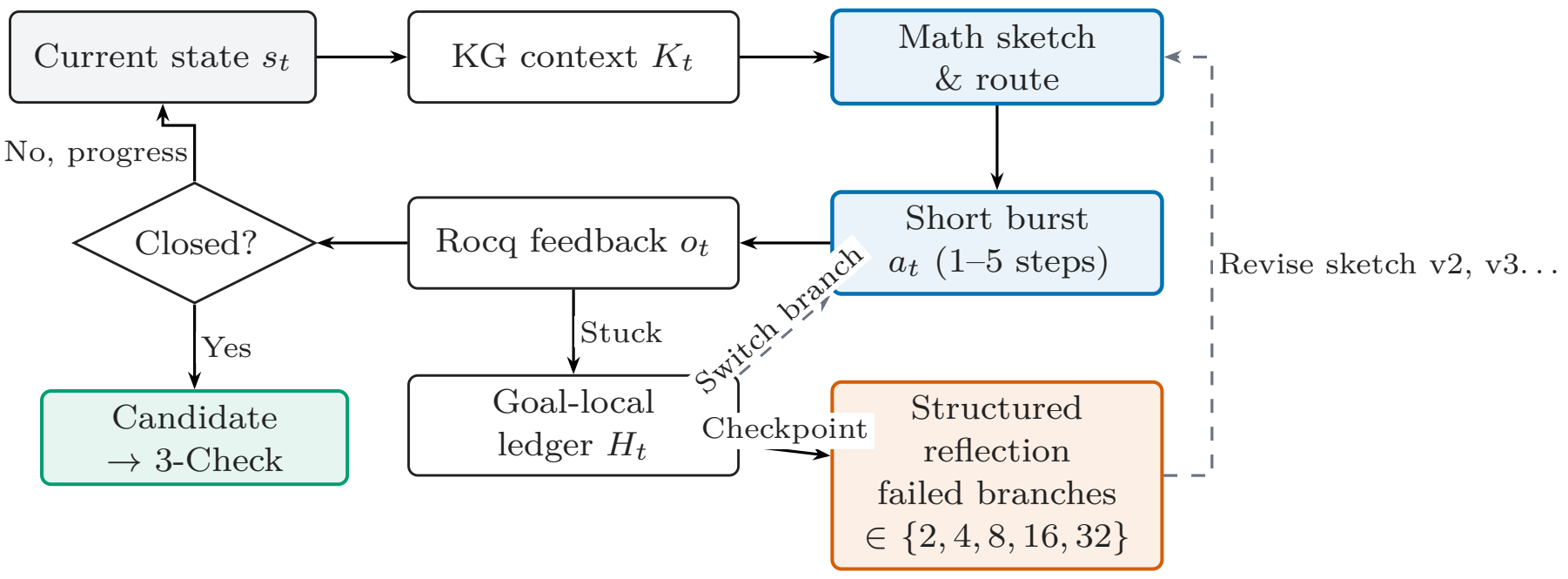
\begin{figure}[htbp]
\centering
\begin{tikzpicture}[node distance=4.5mm and 7.5mm,scale=0.92,transform shape]
\node[databox,text width=23mm] (goal) {Current state $s_t$};
\node[box,text width=25mm,right=of goal] (ret) {KG context $K_t$};
\node[agentbox,text width=25mm,right=of ret] (plan) {Math sketch \& route};
\node[agentbox,text width=25mm,below=7mm of plan] (burst) {Short burst $a_t$ (1--5 steps)};
\node[box,text width=25mm,left=of burst] (rocq) {Rocq feedback $o_t$};
\node[diamond,draw=fmInk,line width=.55pt,aspect=2.0,inner sep=1pt,text width=14mm,align=center,left=of rocq] (done) {Closed?};
\node[okbox,text width=23mm,below=7mm of done] (cand) {Candidate $\to$ 3-Check};
\node[box,text width=25mm,below=7mm of rocq] (ledger) {Goal-local ledger $H_t$};
\node[gate,text width=25mm,below=7mm of burst] (refl) {Structured reflection\\failed branches $\in\{2,4,8,16,32\}$};
\draw[edge] (goal) -- (ret);
\draw[edge] (ret) -- (plan);
\draw[edge] (plan) -- (burst);
\draw[edge] (burst) -- (rocq);
\draw[edge] (rocq) -- (done);
\draw[edge] (done) -- node[lab,right]{Yes} (cand);
\draw[edge] (done.north) |- node[lab,pos=.25,left]{No, progress} ($(goal.south)+(0,-1.8mm)$) -- (goal.south);
\draw[edge] (rocq) -- node[lab,right]{Stuck} (ledger);
\draw[edge] (ledger) -- node[lab,above]{Checkpoint} (refl);
\draw[feedback] (refl.east) -- ++(4mm,0) |- node[lab,pos=.25,right]{Revise sketch v2, v3\ldots} (plan.east);
\draw[feedback] (ledger.north east) -- node[lab,pos=.55,sloped,above]{Switch branch} (burst.south west);
\end{tikzpicture}
\caption{Feedback-driven short-burst proving loop with exponential-interval goal-local reflection.}
\label{fig:loop}
\end{figure}

As shown in Figure~\ref{fig:loop}, for a nontrivial theorem the agent first retrieves KG context $K_t$ and writes a mathematical proof sketch, then interacts with Rocq in \emph{short bursts} of 1--5 tactics $a_t\leftarrow\operatorname{Agent}(s_t,K_t,H_t)$, immediately inspecting the returned state or error $o_t\leftarrow\operatorname{Rocq}(s_t,a_t)$ and using \tool{rocq_step_multi} to probe alternative tactics at choice points without advancing the main state.

We distinguish a failed tactical probe (\emph{burst}) from an abandoned strategy (\emph{branch}). Each abandoned route is logged in the per-goal ledger $H_t$ with its key tactics, last valid state, and blocking subgoal. When failed branches on the current goal reach an exponential checkpoint in $\{2,4,8,16,32\}$, the session triggers \textbf{structured reflection}: it inspects the exact goal, frozen premise packet, and ledger, diagnoses the root cause (mathematical route, formal representation, or tactic usage), writes a revised proof sketch specifying how the next falsifiable burst differs from prior failures, and resumes search automatically.

\subsection{No Cross-Theorem Trajectory Memory and Three-Tier Knowledge Triage}
In early prototypes, we experimented with persisting successful script templates and failure-recovery trajectories across theorems for retrieval into subsequent proofs. Controlled ablations showed that cross-theorem execution trajectories yielded negligible transfer gains on new theorems while adding context overhead and benchmark leakage risks. Megrez therefore keeps the branch ledger \textbf{strictly local to the active goal} and discards it upon completion.

Instead, reusable knowledge is routed through a \textbf{three-tier knowledge triage} discipline governed by attention cost: (1) top-level workflow skills (\tool{SKILL.md}, hard cap $\le 150$ lines), loaded whenever a skill triggers and restricted to workflow invariants; (2) reference idiom manuals (\tool{reference-*.md}, soft cap $\le 150$ lines), loaded on demand for corpus-wide MK and Rocq idioms; and (3) concept-cluster \tool{usage_hints} in the knowledge graph (unbounded capacity), recalled only when a relevant topic cluster is queried. In-proof subgoals (\tool{assert}/\tool{enough}) are used freely, whereas any new standalone helper lemma (including \tool{Local Lemma}) must be stated with its exact type and dependencies and approved at Gate~1.

%======================================================================
\section{Corpus Providers and Recoverable Multi-Theorem Batches}\label{sec:batch}
%======================================================================
\subsection{Provider Stack and Proof-Stripped Dual Fingerprints}
Megrez decouples workflow tooling from mathematical corpora via local \emph{Providers}, each carrying a \tool{provider.json} manifest declaring its version, Rocq version, corpus paths, metadata directories, axiom whitelist, and Provider dependencies. Providers form a directed acyclic stack in which exactly one corpus is mounted as the writable Workspace and all others are read-only dependencies.

For multi-theorem workloads, \tool{FORMALIZATION_BATCH.yaml} is initialized as the single source of truth only after all target statements pass Gate~1 (generating \tool{CHECKLIST.md} automatically). Over an explicit dependency DAG $D=(U,F)$, the ready set is $R=\{v\in U\mid \operatorname{status}(v)=\mathrm{todo},\ \forall(u,v)\in F,\ \operatorname{status}(u)=\mathrm{qed}\}$. To allow parallel worker search across independent theorems in the same \tool{.v} file, the parser computes two SHA-256 fingerprints per target: a normalized \textbf{statement fingerprint} (\tool{statement_sha256}) and a \textbf{proof-stripped context fingerprint} (\tool{context_sha256}) over the preceding file prefix with earlier proof bodies removed. Replacing an earlier theorem's \tool{Admitted} with a completed \tool{Qed} proof in the same file therefore preserves later theorems' context fingerprints without false \tool{stale} invalidation, while any edit to preceding definitions, notations, or target statements is caught immediately.

\subsection{Parallel Read-Only Search, Hash-Bound Integration, and Four-Way Recovery}
\begin{figure}[htbp]
\centering
\begin{tikzpicture}[
  scale=0.90,transform shape,
  st/.style={draw=fmInk,line width=.6pt,rounded corners=6pt,minimum width=19mm,minimum height=7mm,align=center,font=\fontsize{8.5}{10}\selectfont\ttfamily},
  tr/.style={edge},
  tl/.style={lab,font=\fontsize{7}{8.5}\selectfont}
]
\node[st] (todo) at (0,0) {todo};
\node[st] (leased) at (3.4,0) {leased};
\node[st] (cand) at (7.0,0) {candidate};
\node[st] (integ) at (10.8,0) {integrating};
\node[st,draw=fmGreen,fill=fmGreenBg,line width=.85pt] (qed) at (10.8,-2.2) {qed};
\node[st,draw=fmOrange,fill=fmOrangeBg] (blocked) at (3.4,-2.2) {blocked};
\node[st,draw=fmGray,fill=fmGrayBg] (stale) at (7.0,-2.2) {stale};
\draw[tr] (todo) -- node[tl,above]{claim} node[tl,below]{deps qed} (leased);
\draw[tr] (leased) -- node[tl,above]{submit} (cand);
\draw[tr] (cand) -- node[tl,above]{begin\_integ} node[tl,below]{hashes match} (integ);
\draw[tr] (integ) -- node[tl,right,align=left]{4 checks \&\\hash bind pass} (qed);
\draw[tr] (integ.north) to[bend right=25] node[tl,above]{check failed} (cand.north);
\draw[tr] (leased.north) to[bend right=30] node[tl,above]{retry / lease expired} (todo.north);
\draw[tr] (leased) -- node[tl,left,align=right]{checkpoint} (blocked);
\draw[tr] (cand.south west) -- node[tl,sloped,above]{reject \& archive} (blocked.north east);
\draw[tr] (cand) -- node[tl,right]{hash mismatch} (stale);
\draw[tr] (blocked.west) -| node[tl,pos=.3,below]{resume after reflection} (todo.south);
\draw[tr,dashed,draw=fmGray] (stale.south) -- ++(0,-0.45) -| node[tl,pos=.25,below]{human approves source edit} ($(todo.south)+(-0.4,0)$);
\end{tikzpicture}
\caption{Per-theorem state machine in a recoverable batch.}
\label{fig:batch}
\end{figure}
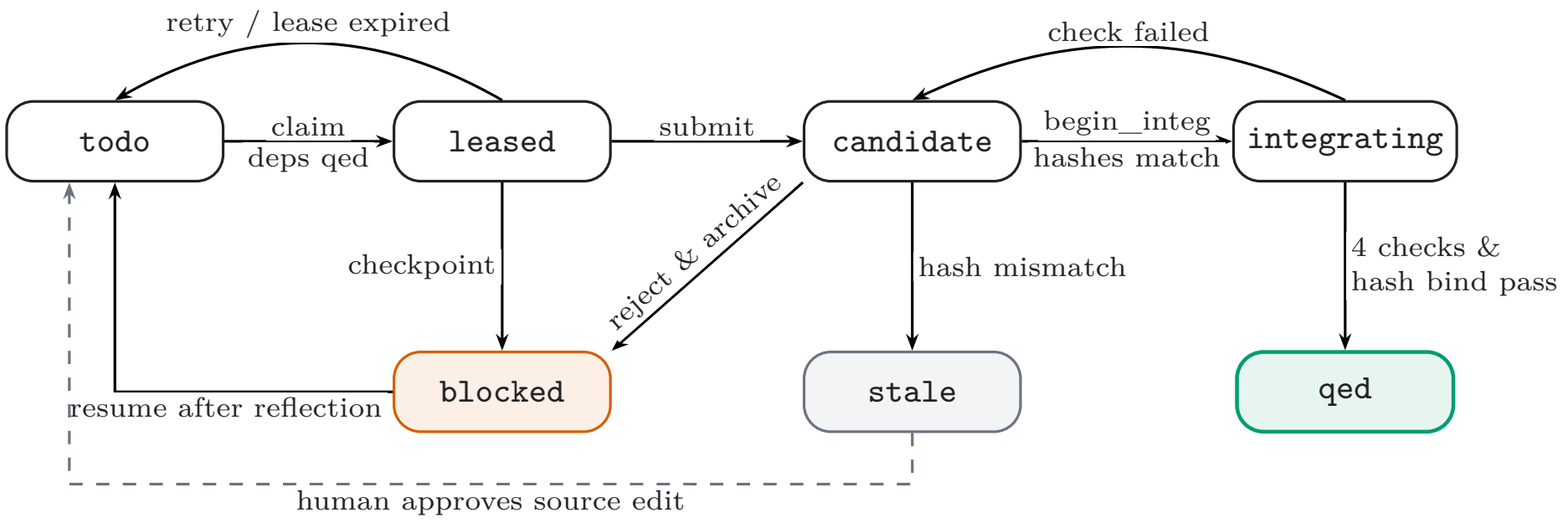

As shown in Figure~\ref{fig:batch}, read-only workers launched with isolated context (\tool{fork_turns="none"}) claim ready goals in parallel (\tool{todo}$\to$\tool{leased}), returning locally verified scripts (\tool{candidate}) or entering \tool{blocked} when attempts cross $\{2,4,8,16,32\}$. The main session holds a private integration capability token (\tool{integration_capability}, stored only as a SHA-256 hash in the manifest; losing it fails closed) and performs serial integration under compare-and-swap revision numbers: it verifies both fingerprints (\tool{integrating}), runs the four acceptance checks (\tool{rocq_verify}, \tool{assumptions_ok}, \tool{file_compile}, \tool{full_build}) plus intra-batch dependency verification, and hash-binds the canonicalized source proof (normalizing comments and whitespace while preserving token boundaries such as \tool{:=}) to the persisted candidate before recording \tool{qed}. If any check fails, the main session reverts the file to \tool{Admitted}, returns to \tool{candidate}, and invokes \tool{batch_reject_candidate} to atomically archive the rejected script and failure evidence into the goal's ledger before transitioning to \tool{blocked}.

After an interruption, sequential execution of \tool{validate}, \tool{recover}, and \tool{status} classifies each interrupted theorem into one of four deterministic recovery actions: (1) \tool{redispatch} (expired \tool{leased} entry with \tool{Admitted} source resets to \tool{todo}); (2) \tool{rewrite_candidate} (\tool{integrating} interrupted before file write returns to \tool{candidate}); (3) \tool{reverify} (\tool{integrating} whose persisted candidate is already written into the file with \tool{Qed} resumes directly at the four acceptance checks); and (4) \tool{manual} (unexpected source drift freezes the entry for human inspection).

%======================================================================
\section{System Evaluation}\label{sec:eval}
%======================================================================
\subsection{MKTopos Corpus and Knowledge Graph Scale}
We deploy Megrez on MKTopos (Rocq~9.1), a formalization of Xiong's \emph{Lectures on Point-Set Topology}~\cite{xiong} built over MK set theory. Tables~\ref{tab:kgstats} and~\ref{tab:chapters} summarize the corpus and its runtime knowledge graph. Across 12 modules/chapters and 262 \tool{.v} files totaling 102,122 lines of proof script, the graph indexes 3,894 formal declarations (including 2,529 completed \tool{Qed} proofs) and 60,982 reference edges, of which 96.8\% (60,795) are resolved precisely from compiler \tool{.glob} files. Curated metadata covers 56 concept clusters (1,678 memberships), 982 aliases, and 415 textbook items across 54 sections (331 formalized).

\begin{table}[htbp]
\centering\small
\caption{Overall scale and structure of the MKTopos knowledge graph.}
\label{tab:kgstats}
\begin{tabularx}{\linewidth}{L{31mm}Y r}
\toprule
Category & Breakdown & Count \\
\midrule
Source scale & \tool{.v} source files / Total lines of completed proof script & 262 / 102,122 \\
\midrule
\multirow{3}{*}{Declarations ($V$)} & Theorem (652) / Lemma (1,718) / Corollary (97) / Prop.\ (59) / Fact (116) & 2,642 \\
 & Definition (903) / Abbreviation (189) / Notation (55) / Other (Instance, Ltac, etc.\ 105) & 1,252 \\
 & Total nodes (with completed \tool{Qed} proofs) & 3,894 (2,529) \\
\midrule
\multirow{3}{*}{Reference edges ($E$)} & \tool{proof_uses} / \tool{statement_mentions} & 32,227 / 12,979 \\
 & \tool{uses_notation} / \tool{uses_tactic} / \tool{file_requires} / \tool{dual_of} \& \tool{notation_of} & 12,455 / 1,745 / 1,493 / 83 \\
 & Total reference edges (resolved from compiler \tool{.glob}) & 60,982 (96.8\%) \\
\midrule
Semantics \& FTS & Concept clusters (members) / Aliases / Textbook sections (items, formalized) / Search terms & 56 (1,678) / 982 / 54 (415, 331) / 23,384 \\
\bottomrule
\end{tabularx}
\end{table}

\begin{table}[htbp]
\centering\small
\caption{Module and chapter breakdown of the MKTopos corpus.}
\label{tab:chapters}
\begin{tabularx}{\linewidth}{lYrrrr}
\toprule
Module / Chapter & Mathematical topic & Files & Decls & Proved & Proof lines \\
\midrule
\tool{MK} & MK set theory (axioms, ordinals, cardinals) & 3 & 437 & 284 & 2,517 \\
\tool{Real_Axioms} + \tool{Real_Analysis} & Real number axioms and basic analysis & 6 & 340 & 227 & 3,543 \\
\tool{Chapter1_NaiveSetTheory} & Set operations, countability, AC \& Tukey & 9 & 149 & 112 & 1,720 \\
\tool{Chapter2_MetricSpaces} & Metric and topological spaces, continuity & 7 & 172 & 46 & 2,453 \\
\tool{Chapter3_SubspacesProducts} & Subspaces, finite/arbitrary products, quotients & 4 & 117 & 47 & 1,628 \\
\tool{Chapter4_Connectedness} & Connectedness, components, path/local conn. & 5 & 148 & 111 & 5,555 \\
\tool{Chapter5_CountabilityAxioms} & First/second countability, Lindel\"of, separable & 4 & 91 & 64 & 2,643 \\
\tool{Chapter6_SeparationAxioms} & Separation axioms, Urysohn, Tietze, metrization & 7 & 356 & 270 & 10,048 \\
\tool{Chapter7_Compactness} & Compactness, Tychonoff, local compactness & 8 & 93 & 62 & 8,468 \\
\tool{Chapter8_CompleteMetricSpaces} & Complete metric spaces, Baire category, completion & 2 & 161 & 132 & 6,324 \\
\tool{Chapter9_MappingSpaces} & Mapping spaces, compact-open, Arzel\`a--Ascoli & 3 & 35 & 14 & 3,025 \\
\tool{Chapter10_FundamentalGroup} & $\pi_1(S^1)$, covering maps, Brouwer \& Jordan thms & 204 & 1,795 & 1,160 & 54,198 \\
\midrule
Total & & 262 & 3,894 & 2,529 & 102,122 \\
\bottomrule
\end{tabularx}
\end{table}

\subsection{Retrieval Latency and Scalability Benchmarks}
To evaluate scalability as libraries grow by 1--2 orders of magnitude, we use a deterministic synthetic generator (seed \tool{20260901}, matching the corpus distribution with 5\% Chinese comments, 4\% formal operator tokens, 10\% node aliases, and 0.1\% naming collisions) to construct 10,000-node ($2.5\times$ MKTopos, 16.9\,MB, built in 1.05\,s) and 100,000-node ($25\times$ MKTopos, 170.4\,MB, built in 10.84\,s) databases. Table~\ref{tab:perf} reports 95th-percentile latency (p95 over 7 repetitions on Apple Silicon arm64, 10 cores, Python 3.12.11, SQLite 3.50.3) and executed SQL statements per call.

\begin{table}[htbp]
\centering\small
\caption{Interactive and premise retrieval performance on 10k- and 100k-node graphs (p95 latency in ms; SQL statement count in parentheses).}
\label{tab:perf}
\begin{tabularx}{\linewidth}{lYrrrr}
\toprule
Task \& workload & Query example & \multicolumn{2}{c}{10,000 nodes (16.9\,MB)} & \multicolumn{2}{c}{100,000 nodes (170.4\,MB)} \\
\cmidrule(lr){3-4}\cmidrule(lr){5-6}
 & & \tool{legacy-v1} & \tool{fts-v1} & \tool{legacy-v1} & \tool{fts-v1} \\
\midrule
Interactive: Exact identifier & \tool{Synthetic_000001} & 8.11 (7--24) & \textbf{0.71} (3) & 72.29 (7--24) & \textbf{1.23} (3) \\
Interactive: Chinese concept & \texttt{conn.\ space (ZH)} & 8.30 (16) & \textbf{6.21} (7) & 76.87 (16) & \textbf{19.73} (7) \\
Interactive: Formal operators & \tool{/\ -> :=} & 6.33 (6) & \textbf{6.12} (7) & 58.51 (6) & \textbf{18.10} (7) \\
Interactive: Common token & \tool{topology common} & 40.11 (36) & \textbf{24.33} (7) & 410.80 (36) & \textbf{199.24} (7) \\
Interactive: Single CJK char & \texttt{conn.\ char (ZH)} & 8.99 (36) & \textbf{0.91} (5) & 82.55 (36) & \textbf{0.99} (5) \\
\midrule
Premise: Exact identifier & \tool{Synthetic_000001} & 3.02 (3) & 25.42 (6) & 20.54 (3) & 213.98 (6) \\
Premise: Chinese semantic & \texttt{conn.\ space (ZH)} & 2.92 (2) & 9.06 (5) & 19.83 (2) & 41.92 (5) \\
Premise: Formal operators & \tool{/\ -> :=} & 2.89 (2) & 8.88 (5) & 20.04 (2) & 40.38 (5) \\
Premise: Common token & \tool{topology common} & 3.09 (3) & 26.26 (6) & 20.34 (3) & 217.18 (6) \\
\bottomrule
\end{tabularx}
\end{table}

Two findings stand out. First, in \textbf{interactive search} (\tool{kg_search}), \tool{fts-interactive-v1} outperforms \tool{legacy-search-v1} on every workload: at 100k nodes, exact lookup accelerates by \textbf{58.8$\times$} (72.29\,ms $\to$ 1.23\,ms), Chinese concept search by \textbf{3.9$\times$}, and common-token search by \textbf{2.1$\times$}, while reducing SQL statements from up to 36 to a bounded 3--7 per call and recalling 10 hits on pure operator queries (\tool{/\ -> :=}) where legacy search returns 0. Second, in \textbf{premise retrieval} (\tool{kg_context_packet}), \tool{legacy-premise-v1} performs only identifier and passed-concept lookups without full-text matching, whereas \tool{fts-premise-v1} materializes a request-private \tool{TEMP eligible_nodes} table for scope-before-cap filtering and fuses seven-column BM25 with graph neighborhoods via RRF in under 27\,ms at 10k nodes and 40--217\,ms at 100k nodes.

\subsection{Batch Scheduling and Engineering Contract Verification}
Megrez includes 39 test modules with 319 automated contract and fault-injection tests (Table~\ref{tab:tests}), all passing in continuous integration. In end-to-end multi-theorem scenario tests (\tool{test_batch_scenario.py}) over dependent theorems $A,B\to C$ in a single \tool{.v} file, two read-only workers claim $A$ and $B$ in parallel; injecting crashes before and after source writing on $B$ deterministically triggers \tool{rewrite_candidate} and \tool{reverify}; completing $B$'s proof preserves $A$'s proof-stripped context fingerprint without false \tool{stale} invalidation; and $C$ unlocks automatically once both $A$ and $B$ reach \tool{qed}, completing after simulated lease expiration and \tool{redispatch}.

\begin{table}[htbp]
\centering\small
\caption{Product-level engineering contract and fault-injection test suite (39 files, 319 tests).}
\label{tab:tests}
\begin{tabularx}{\linewidth}{L{34mm}rrY}
\toprule
Test category & Files & Tests & Verified invariants and fault-injection scenarios \\
\midrule
Source \& ref parsing & 5 & 43 & \tool{.v} declaration/proof boundary parsing, terminator detection after focused blocks, \tool{.glob} extraction, proof-stripped context fingerprints \\
KG build \& Providers & 10 & 80 & Idempotent rebuilds, fault injection with atomic \tool{.previous} rollback, dual-schema support, Provider topological sort, single writable Workspace \\
Retrieval \& scope safety & 11 & 69 & FTS5 injection-safe compilation, pinned navigation + RRF ranking, scope-before-cap filtering, self-reference masking, token budgeting \\
Batch \& crash recovery & 7 & 120 & DAG readiness, CAS revisions, capability-token auth, same-file parallel claims, 4-way recovery (\tool{redispatch}/\tool{rewrite_candidate}/\tool{reverify}/\tool{manual}), candidate rejection \\
Workflow \& packaging & 6 & 7 & Attention-budget line caps ($\le 150$ lines), cooperative read-only worker contracts, Gate~1 review before batch init, zero trajectory memory \\
\bottomrule
\end{tabularx}
\end{table}

%======================================================================
\section{Collaborative Formalization Case Studies}\label{sec:case}
%======================================================================
\subsection{Case Study 1: Axiom of Choice and Tukey's Lemma (Textbook \S1.8)}
Section~1.8 of Xiong's textbook~\cite{xiong} covers 4 definitions and 5 theorems/lemmas/corollaries (Table~\ref{tab:sec18}), illustrating concept alignment and reflection-driven lemma decomposition:
\begin{itemize}
  \item \textbf{Concept alignment (\tool{mk-formalize}):} Running \tool{kg_textbook("1.8")} and \tool{kg_search} uncovered two structural alignments: Definition~1.8.2 (``nest of a family'') directly matched \tool{Nest} in the foundational MK library, avoiding duplicate definitions; and Axiom~1.8.1 (``every set admits a choice function'') became a provable theorem \tool{Xiong_1_8_1} (12 lines) because MK's global choice axiom \tool{AxiomIX} already supplies a choice function on the universe $\mu\setminus\{\Phi\}$, which need only be restricted to $\mathcal P(X)\setminus\{\Phi\}$.
  \item \textbf{Reflection-driven helper lemmas (\tool{mk-prove}):} Proving Lemma~1.8.3 (\tool{Xiong_1_8_3}, closure of finite-character families under nest unions) requires showing that every finite subset $B\subset\bigcup\mathcal C$ lies in a single member of the nest $\mathcal C$. Initial bursts attempting direct induction on $B$ failed because finiteness in MK is defined cardinally as $P[B]\in\omega$ (\tool{Finite}) and applying \tool{Mathematical_Induction} directly with a quantified set variable $B$ fails higher-order unification. At the reflection checkpoint, the main session factored the proof into two standalone lemmas: \tool{remove_one_card} (16 lines, using \tool{Lemma163b}, \tool{finue}, and successor injectivity \tool{MKT136} to reduce cardinality from $n^+$ to $n$) and \tool{fin_chain_union} (40 lines, passing an explicit predicate \tool{apply (Mathematical_Induction (fun n => ...))} and merging nest members by total ordering). This closed \tool{Xiong_1_8_3} in 14 lines. For Tukey's Lemma (\tool{Xiong_1_8_4}), KG search on ``maximal'' surfaced Hausdorff's maximal principle \tool{MKT143}, yielding a 32-line proof via a maximal nest $\mathcal N\subset\mathcal F$ instead of the textbook's five-step ``$t$-subfamily'' construction.
  \item \textbf{Knowledge sedimentation (\tool{mk-kg-update}):} Following full compilation, the update stage registered the concept cluster \tool{xiong_choice_tukey} and recorded the global-choice restriction and explicit-predicate cardinal induction idioms into \tool{usage_hints} and the idiom manual.
\end{itemize}

\begin{table}[htbp]
\centering\small
\caption{Formalization summary for Textbook \S1.8 in \tool{Sec1_8_AxiomOfChoiceTukey.v}.}
\label{tab:sec18}
\begin{tabularx}{\linewidth}{lL{31mm}lrY}
\toprule
Textbook item & Formal identifier & Kind & Lines & Key design or proof route \\
\midrule
Defs.\ 1.8.1--1.8.4 & \tool{SelectionFun}, \tool{FiniteChar}, \tool{MaxMember} / \allowbreak\tool{MinMember} & \tool{definition} & -- & Local choice function, finite-character family, maximal/minimal members; Def.\ 1.8.2 reuses MK base \tool{Nest} \\
Axiom 1.8.1 / Thm.\ 1.8.2 & \tool{Xiong_1_8_1} / \tool{Xiong_1_8_2} & \tool{theorem} & 12 / 19 & Restricts global choice \tool{AxiomIX} to $\mathcal P(X)\setminus\{\Phi\}$ and indexed family $A$ via \tool{MKT126a/b/c} and \tool{MKT30} \\
(Helper lemmas) & \tool{remove_one_card} / \tool{fin_chain_union} & \tool{lemma} & 16 / 40 & Removing one element reduces $P[B]=n^+$ to $n$; explicit-predicate induction on $n=P[B]$ over nests \\
Lemma 1.8.3 & \tool{Xiong_1_8_3} & \tool{theorem} & 14 & Nonempty finite-character families are closed under subsets and nest unions (via \tool{fin_chain_union}) \\
Thm.\ 1.8.4 / Cor.\ 1.8.5 & \tool{Xiong_1_8_4} / \tool{Xiong_1_8_5} & \tool{thm} / \tool{cor} & 32 / 63 & Tukey's Lemma via Hausdorff's principle \tool{MKT143} on a maximal nest; Corollary on $\mathcal F_A=\{B\in\mathcal F\mid A\subset B\}$ \\
\bottomrule
\end{tabularx}
\end{table}

\subsection{Case Study 2: Connectedness from Finite to Arbitrary Products (Chapter 4)}
Chapter~4 (Connectedness; 148 declarations, 111 proofs, 5,555 lines) demonstrates cross-section reuse via the knowledge graph:
\begin{enumerate}
  \item \textbf{Coordinate-update chains on indexed products (\S4.1):} In MK set theory, binary Cartesian products $X\times Y$ (ordered pairs) and indexed products $\prod_{\gamma\in G}X_\gamma$ (\tool{IProduct}, choice functions on $G$) have distinct encodings. Rather than inducting on binary products and transporting across homeomorphisms, the agent worked directly on \tool{IProduct}: it defined single-coordinate updates (\tool{CoordinateUpdateMap}), proved their images are connected slices (41 lines), and inducted over finite index sets $G$ to link slices into \tool{FiniteCoordinateUpdateChain} (87 lines), proving finite-product connectedness (\tool{Xiong_4_1_9}) in 36 lines.
  \item \textbf{Cross-section reuse I: punctured $\mathbb R^n$ connectedness and $\mathbb R^n\not\cong\mathbb R$ (\S4.2):} Whereas the textbook proves only for $n=2$ that $\mathbb R^2\setminus\{0\}$ is connected and $\mathbb R^2\not\cong\mathbb R$, querying \tool{kg_concepts} for \tool{xiong_connectedness} retrieved the \S4.1 coordinate slices. Adding a 13-line lemma (\texttt{\small CoordinateUpdate\allowbreak Slice\_\allowbreak avoids\_point}, showing that updating coordinate $\alpha$ avoids the origin whenever some other coordinate $\beta\neq\alpha$ is nonzero) generalized punctured connectedness and $\mathbb R^n\not\cong\mathbb R$ to all finite dimensions $n\ge 2$ (\tool{Xiong_4_2_6}, 129 lines; \tool{Xiong_4_2_7}, 26 lines).
  \item \textbf{Cross-section reuse II: arbitrary product connectedness (\S4.3):} In \S4.3, querying \tool{kg_deps} on \tool{FiniteCoordinateUpdateChain} provided the template for Tychonoff's connectedness theorem: generalizing coordinate updates to finite subsets $F\subset G$ (\tool{FiniteGeneralCoordinateUpdateChain}, 72 lines) showed that the connected component $C$ of a basepoint meets every basic open box and is dense; combined with component closure (\tool{Xiong_4_3_1}), this proved arbitrary product connectedness (\tool{Xiong_4_3_2}) in 59 lines.
\end{enumerate}

\subsection{Case Study 3: 2D Brouwer Fixed-Point and Jordan Separation Theorems (Chapter 10)}
In Chapter~10 (204 files, 1,795 declarations, 54,198 lines), two mechanisms proved decisive: (1) in \S10.5 (2D no-retraction \tool{Xiong_10_5_1} and Brouwer fixed-point theorem \tool{Xiong_10_5_2}), reflection replaced pointwise algebraic composition on $\mathbb R$ with subspace domain restriction (restricting square roots to nonnegative reals and inversion to positive reals, composed via subspace corestriction), eliminating nonnegativity and nonzero-denominator obligations from continuity proofs; and (2) in \S10.6 (Jordan separation theorem \tool{Xiong_10_6_6}), decomposing non-separation by arcs (\tool{Xiong_10_6_1}) and disks (\tool{Xiong_10_6_2}), boundary properties (\tool{Xiong_10_6_4}), and the square-crossing lemma (\tool{Xiong_10_6_5}) into modular batches enabled parallel worker search followed by serial main-session integration into \tool{xiong_jordan_curve}.

%======================================================================
\section{Related Work, Limitations, and Conclusion}\label{sec:related}
%======================================================================
\paragraph{Related work.} Neural theorem provers have evolved from state-conditioned step generation~\cite{gptf,leandojo}, informal-to-formal sketching~\cite{dsp}, and in-context proving agents~\cite{copra} to reinforcement-learned whole-proof models~\cite{deepseekprover2,kimina,goedel2} and general coding agents coupled with MCP tools such as Numina-Lean-Agent~\cite{numina}. In Coq/Rocq, classical automation includes CoqHammer~\cite{coqhammer}, Tactician~\cite{tactician}, and Proverbot9001~\cite{proverbot}, while recent LLM systems include PALM~\cite{palm}, CoqPilot~\cite{coqpilot}, Rango~\cite{rango}, and Baldur~\cite{baldur} in Isabelle. Premise selection and agent reflection have been studied via dense retrievers~\cite{leandojo,magnushammer} and verbal reflection loops~\cite{react,reflexion}. Megrez complements these efforts by targeting cumulative textbook formalization over low-resource MK set theory, replacing black-box embeddings with an explicit, source-rebuildable knowledge graph and integrating short-burst proving, goal-local reflection, three-check acceptance, and recoverable batches.

\paragraph{Limitations and discussion.} (1) \textbf{Metadata and edge semantics:} Retrieval quality depends on curated concepts and Chinese aliases, though reporting every exploratory \tool{Search} fallback hit exposes missing annotations organically; compiler \tool{.glob} edges record symbol occurrences rather than minimal logical dependencies, so soundness always rests on the Rocq kernel and \tool{rocq_assumptions}. (2) \textbf{Candidate scope boundaries:} \tool{CandidateScopeSnapshot} enforces node-set membership prior to SQL lane caps rather than full Rocq module/notation visibility, and full-text IDF statistics span all configured corpora; audit envelopes therefore explicitly record \tool{scope_authority} and \tool{all_configured}. (3) \textbf{Trust boundaries:} CAS revisions, capability hashes, and canonical proof-hash bindings prevent concurrent drift and proof swapping, whereas cooperative read-only workers share a workspace and do not replace OS-level sandboxing for untrusted processes.

\paragraph{Conclusion.} We presented Megrez, a knowledge-guided formalization agent for Rocq libraries over Morse--Kelley set theory. By combining a source-rebuildable knowledge graph, a 1--5 tactic short-burst proving loop with exponential-interval goal-local reflection, and a three-check acceptance pipeline with recoverable multi-theorem batches, Megrez supports scalable retrieval and cross-chapter formalization across the 3,894 declarations and 102,122 proof lines of the MKTopos corpus, demonstrating a practical engineering path toward textbook-scale formal mathematics.

\paragraph{Code and corpus availability.} The Megrez formalization agent and knowledge-graph toolchain are publicly available at \url{https://github.com/dgjszy/Megrez-Rocq-Agent}, and the accompanying point-set topology formalization corpus over Morse--Kelley set theory is available at \url{https://github.com/dgjszy/RocqAtlas}.

\paragraph{Acknowledgments.} GPT 5.6 was used during this research to assist with system development, formalization exploration, and manuscript preparation.

%======================================================================

\end{document}